%% file: AnalysisCIsystems.tex
\documentclass[10pt,journal,cspaper,compsoc]{IEEEtran}
\pdfoutput=1
\usepackage{times}
\usepackage{epsfig}
\usepackage{graphicx}
\usepackage{amsmath}
\usepackage{amssymb}
\usepackage{color}

\usepackage[pagebackref=true,breaklinks=true,letterpaper=true,colorlinks,bookmarks=false]{hyperref}

\newcommand{\Section}[1]{\vspace{-3pt}\section{#1}\vspace{-3pt}}
\newcommand{\Subsection}[1]{\vspace{-2pt}\subsection{#1}\vspace{-2pt}}

\newenvironment{tight_itemize}{\begin{itemize} \itemsep
-1pt}{\end{itemize}}
\newenvironment{tight_enumerate}{\begin{enumerate} \itemsep
-1pt}{\end{enumerate}}

\begin{document}

\title{\huge{A Framework for Analysis of Computational Imaging Systems: Role of Signal Prior, Sensor Noise and Multiplexing}}

\author{Kaushik~Mitra,~\IEEEmembership{Member,~IEEE}, Oliver~S.~Cossairt,~\IEEEmembership{Member,~IEEE}
        and~ Ashok~Veeraraghavan,~\IEEEmembership{Member,~IEEE}
\IEEEcompsocitemizethanks{\IEEEcompsocthanksitem K. Mitra and A. Veeraraghavan are with the Department
of Electrical and Computer Engineering, Rice University, Houston, TX, 77025.\protect\\
E-mail: Kaushik.Mitra@rice.edu and vashok@rice.edu
\IEEEcompsocthanksitem O. S. Cossairt is with the Department
of Electrical Engineering and Computer Science, Northwestern University, Evanston, IL 60208. \protect\\
E-mail: ollie@eecs.northwestern.edu}
\thanks{}
}


\maketitle
\begin{abstract}
\input{abstract.tex}

\end{abstract}


%
\Section{Introduction}
\label{sec:Intro}
\input{intro.tex}

\Section{Related Work}
\label{sec:relatedWork}
\input{relatedWork.tex}

\Section{Problem Definition and Notation}
\label{sec:ProbDef}
\input{ProbDef.tex}

\Section{Analytic Performance Characterization of CI Systems Using GMM Prior}
\label{sec:GMMPrior}
\input{GMMPrior.tex}

\Section{Common Framework for Analysis of CI Systems}
\label{sec:frmAnalCIsystems}
\input{frmAnalCIsystems.tex}

\Section{Performance Analysis of EDOF Systems}
\label{sec:perfEDOFsystems}
\input{perfEDOFsystems.tex}

\Section{Performance Analysis of Motion Deblurring Systems}
\label{sec:perfMotDeblurSystems}
\input{perfMotDeblurSystems.tex}

\Section{Performance Analysis of Light Field Systems}
\label{sec:perfLF}
\input{perfLF.tex}

\Section{Exact MMSE vs. Its Lower and Upper Bounds}
\label{sec:exactVsBounds}
\input{exactMMSEvsBounds.tex}


\Section{Discussions}
\label{sec:Discussions}
\input{Discussions.tex}

\Section{Acknowledgements}
Kaushik Mitra and Ashok Veeraraghavan acknowledge support through NSF Grants NSF-IIS:
1116718, NSF-CCF:1117939 and a research grant from Samsung Advanced Institute of Technology through the Samsung GRO program.


\vspace{-0.1in}
{\small
\bibliographystyle{ieee}
\bibliography{egbib}
}

\input{authorBio.tex}

\end{document}

%% file: abstract.tex
Over the last decade, a number of Computational Imaging (CI) systems have been proposed for tasks 
such as  motion deblurring, defocus deblurring and multispectral imaging.
These techniques increase the amount of light reaching the sensor via multiplexing and then undo the deleterious effects of multiplexing by appropriate reconstruction algorithms.
Given the widespread appeal and the considerable enthusiasm generated by these techniques, a detailed performance analysis of the benefits conferred by this approach is important.

Unfortunately, a detailed analysis of CI has proven to be a challenging problem because performance 
depends equally on three components: (1) the optical multiplexing, (2) the noise characteristics 
of the sensor, and (3) the reconstruction algorithm which typically uses signal priors. 
A few recent papers \cite{cossairt2013does,ratner2007illumination,ihrke2010theory} have performed analysis taking multiplexing and noise characteristics into account. 
However, analysis of CI systems under state-of-the-art reconstruction algorithms, most of which exploit signal prior models, has proven to be unwieldy. 
In this paper, we present a comprehensive analysis framework incorporating all three components. 

In order to perform this analysis, we model the signal priors using a Gaussian Mixture Model (GMM). A GMM prior confers two unique characteristics. Firstly,  GMM satisfies the universal approximation property which says that any prior density function can be approximated to any fidelity using a GMM with appropriate number of mixtures.
Secondly, a GMM prior lends itself to analytical tractability allowing us to derive simple expressions for the `minimum mean square error' (MMSE) which we use as a metric to characterize the performance of CI systems.
We use our framework to analyze several previously proposed CI techniques (focal sweep, flutter shutter, parabolic exposure, etc.), giving conclusive answer to the question:
`How much performance gain is due to use of a signal prior and how much is due to multiplexing? 
Our analysis also clearly shows that multiplexing provides significant performance gains above and beyond the gains obtained due to use of signal priors.

%% file: intro.tex
\begin{figure}[tbh] \centering
\includegraphics[width=\columnwidth]{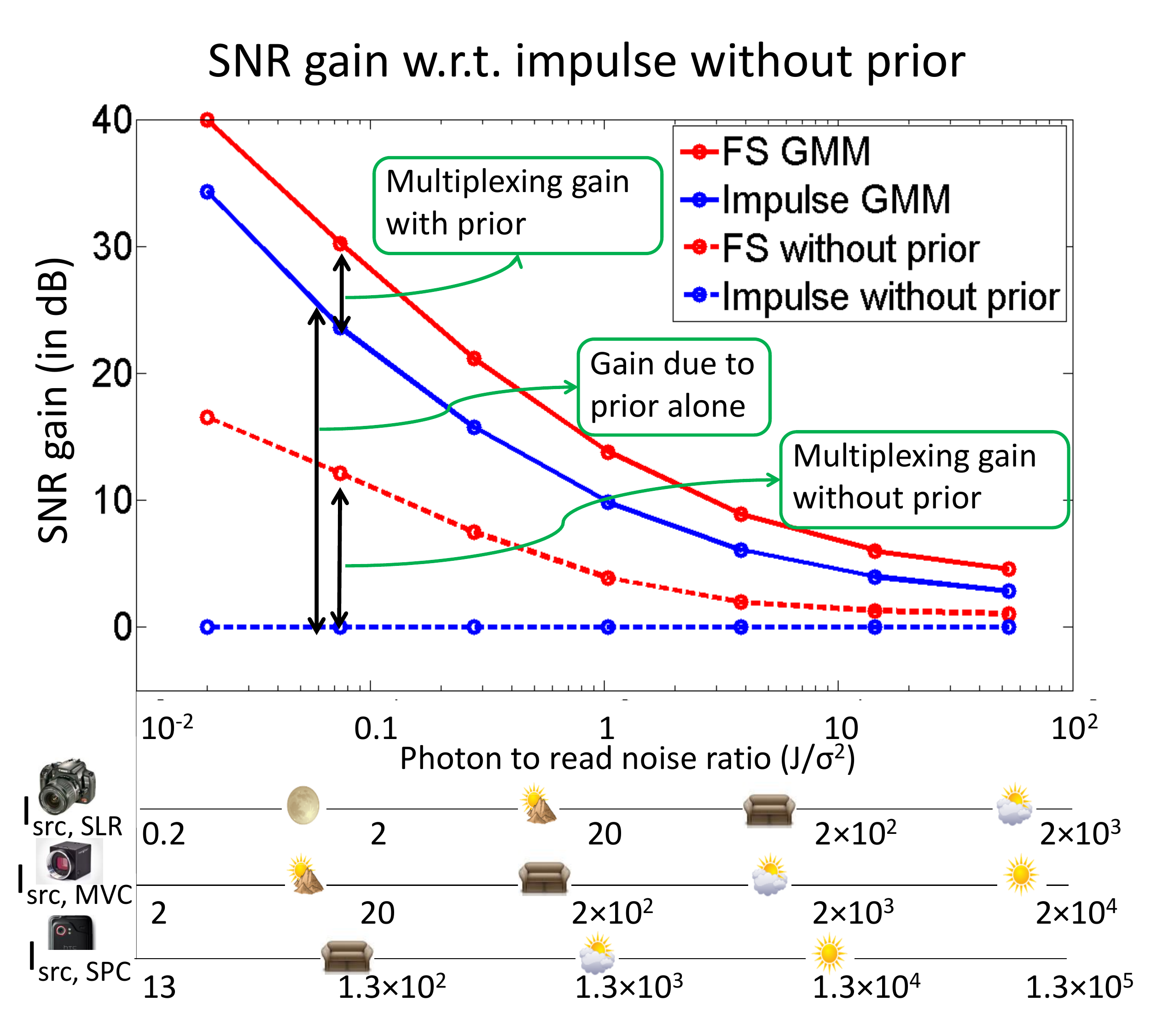} 
\caption{\emph{Effect of signal prior on multiplexing gain of focal sweep \cite{Nagahara:08}:} We show the multiplexing gain of focal sweep over impulse imaging (a conventional camera with stopped down aperture) at different photon to read noise ratios $J/\sigma_r^2$. The photon to read noise ratio is related to illumination level and camera specifications. In the extended x-axis, corresponding to different values of $J/\sigma_r^2$, we show the light levels (in lux) for three camera types: a high end SLR, a machine vision camera (MVC) and a smartphone camera (SPC). As shown by Cossairt et al. \cite{cossairt2013does}, without using signal priors, we get a huge multiplexing gain at low $J/\sigma_r^2$. However, given that most state-of-the-art reconstruction algorithms are based on signal priors, such huge gains are unrealistic. In practice, with the use of signal prior, we get much more modest gains. Our goal is to analyze the multiplexing gain of CI systems above and beyond the use of signal priors. 
}
\label{fig:motiveFig} 
\end{figure}

Computational Imaging systems can be broadly categorized into two categories \cite{nayar2011computational}: 
those designed either to add a new functionality or to increase performance relative to a
conventional imaging system. A light field camera \cite{ng2005light,MaskPaper,lanman2008shield,liang2008programmable} 
is an example of the former: it can be 
used to refocus or change perspective \textit{after} images are captured - a functionality
impossible to achieve with a conventional camera. The latter type of systems are the 
focus of this paper, and from here on we use the term CI to refer to them. 
Examples include extended depth-of-field (EDOF) 
systems \cite{levin2007image,MaskPaper,Zhou:09,Hausler:72,Nagahara:08,dowski1995extended,Hasinoff08light-efficientphotography,latticefocal,georgeLogAsphere,castroAsymPhaseMasks,Ojeda-Castaneda:05,guerrero:07,schroff,fosera}, 
motion deblurring \cite{raskar2006coded,levin2008motion,cho2010motion}, 
spectroscopy \cite{harwit1979hadamard,hanley1999spectral,wagadarikar2008single}, 
color imaging \cite{baer1999comparison,ihrke2010theory}, multiplexed light field acquisition \cite{MaskPaper,lanman2008shield,liang2008programmable,reinterpretableImg}, temporal multiplexing \cite{CodedStrobing,P2C2,hitomi2010video,tempSuperAgrawal,fsvcHolloway} 
and illumination multiplexing \cite{schechner2007multiplexing,ratner2007illumination}. 
These systems use optical coding (multiplexing) to increase light throughput, which increases the 
SNR of captured images. The desired signal is then recovered computationally via signal processing. 
The quality of recovered images depends jointly on the type of optical coding and
the increased light throughput. A poor choice of multiplexing will reduce image quality.

The question of exactly how much performance improvement can be achieved via 
multiplexing has received a fair amount of attention in the literature 
\cite{harwit1979hadamard,CAVE_0316,cossairt2013does,ihrke2010theory,schechner2007multiplexing,wuttig,ratner2007illumination,Hasinoff08light-efficientphotography,Hasinoff:09}.
It is well understood that multiplexing gives the greatest advantage at low light levels 
(where \textit{signal-independent} read noise dominates), but this advantage diminishes with increasing light 
(where \textit{signal-dependent} photon noise dominates) \cite{harwit1979hadamard}. 
However, it is impractical to study the effects of multiplexing alone, since 
signal priors are at the heart of every state-of-the-art reconstruction 
algorithm (e.g. dictionary learning \cite{aharon2006rm}, BM3D \cite{dabov2007image}, GMM \cite{sapiro,mitra2012light}). 
Signal priors can dramatically increase 
performance in problems of deblurring (multiplexed sensing) and denoising (no 
multiplexing), typically with greater improvement as noise increases (i.e. as the light level decreases). 
While both signal priors and multiplexing increase performance at low light levels, 
the former is trivial to incorporate and the latter often requires 
hardware modifications. Thus, it is imperative to understand 
the improvement due to multiplexing \textit{beyond the use of signal priors}.
However, comprehensive analysis of CI systems remains an elusive problem because state-of-the-art
priors often use signal models unfavorable to analysis.     

In this work, we follow a line of research whose goal is to derive bounds on the
performance of CI systems \cite{ratner2007illumination,ihrke2010theory}, and
relate maximum performance to practical considerations (e.g. illumination
conditions and sensor characteristics) \cite{cossairt2013does}. We follow the
convention adopted in
\cite{cossairt2013does,ratner2007illumination,ihrke2010theory}, where the
performance of the CI systems are compared against their corresponding
\emph{impulse imaging systems}, which are defined as a conventional camera that
directly measures the desired signal (e.g. without blur). Noise is related to
the lighting level, scene properties and sensor characteristics. In this paper,
we pay special attention to the problems of defocus and motion blurs and to the
problem of light field acquisition. Defocus and motion blurs can be position
dependent when objects in the scene span either a range of depths or velocities.
Various techniques have been devised to encode blur so as to make it either
well-conditioned or position-independent (shift-invariant), or both. 
 For defocus deblurring, CI systems encode defocus blur using attenuation masks
 \cite{levin2007image,MaskPaper,Zhou:09}, refractive masks
 \cite{dowski1995extended}, or motion \cite{Hausler:72,Nagahara:08}. The impulse
 imaging counterpart is a narrow aperture image with no defocus blur. For motion
 deblurring, CI systems encode motion blur using a fluttered shutter
 \cite{raskar2006coded} or camera motion \cite{levin2008motion,cho2010motion}.
 The impulse imaging counterpart is an image with short exposure time and no
 motion blur. Many camera designs have been proposed to capture light fields
 such as: the microlens array based light field camera (Lytro)
 \cite{ng2005light}, coded aperture camera \cite{liang2008programmable},
 mask-near-sensor designs \cite{MaskPaper,lanman2008shield} and camera array
 designs \cite{camArrayLevoy}. In this paper we analyze only single sensor,
 snapshot light field camera systems. The corresponding impulse camera, which
 directly captures the light field, is a pinhole array mask placed near the sensor. 
 
 Cossairt et al. \cite{cossairt2013does} derived an upper bound stating that the
 maximum gain due to multiplexing is quite large at low light levels. For
 example, in Figure \ref{fig:motiveFig}, the multiplexing gain of focal sweep is
 $> 10$ dB for a low photon to read noise ratio $<0.1$. However, as we show in
 this paper, this makes for an exceptionally weak bound because signal priors
 are not taken into account. In practice, signal priors can be used to improve the performance of any camera,
impulse and computational alike. Since incorporating a signal prior can be done
merely by applying an algorithm to captured images, it is natural to expect that
we would always choose to do so. However, it has historically been very
difficult to determine exactly how much of an increase in performance to expect
from signal priors, making it difficult to provide a fair comparison between
different cameras. 

We present a comprehensive framework that allows us to analyze the performance of
CI systems while simultaneously taking into account multiplexing, sensor noise,
and signal priors. We characterize the performance of CI systems under a GMM
prior which has two unique properties: Firstly, GMM satisfies the universal
approximation property which says that any probability density function (with a
finite number of discontinuities) can be approximated to any fidelity using a
GMM with an appropriate number of mixtures \cite{GMMapproxAllPrior1,
GMMapproxAllPrior2}. Secondly, a GMM prior lends itself to analytical
tractability allowing us to derive simple expressions for the MMSE, which we use
as a metric to characterize the performance of both impulse and computational
imaging systems. We use our framework to analyze several previously proposed CI
techniques (focal sweep, flutter shutter, parabolic exposure, etc.), giving
conclusive answers to the questions: `How much gain is due to the use of a
signal prior and how much is due to multiplexing? What is the multiplexing gain
beyond the use of signal prior?'.

 
 We show that the SNR benefits due to the use of a signal prior alone are quite
large in low light and decrease as the light level increases (see Figure
\ref{fig:motiveFig}). Furthermore, show that when priors are taken into account, multiplexing provides us 
realistic gains of $9.6$ dB for
 EDOF in low light conditions (see Figure \ref{fig:compareEDOFsystems}), $7.5$
 dB for motion deblurring systems in low light conditions (see Figure
 \ref{fig:compareMotDeblurSystems}), and $12$ dB for light field systems in high
 light conditions (see Figure \ref{fig:analSnrGainLF}). These are substantial
 gains as these implies that the MSE of the CI systems are less than those of
 corresponding impulse systems by factors of $9$, $5.5$ and $16$, respectively.
 This indicates that CI techniques improve the performance of traditional
 imaging beyond the benefits conferred due to sophisticated
 reconstruction algorithms.

\Subsection{Key Contributions}
\begin{tight_enumerate}

\item{We introduce a framework for analysis of CI systems under signal priors. Our  analysis is based on the GMM prior, which can approximate almost any probability density function and is analytically tractable.}

\item{We use the GMM prior to quantify exactly how much the use of signal priors can improve the performance of a given camera. We also quantify the multiplexing gain beyond that due to the use of signal priors.}

\item{We analyze the performance of many CI systems \emph{with signal priors taken into account}. We show that the SNR gain due to multiplexing beyond the use of signal priors can be significant ($9.6$ dB for defocus deblurring cameras, $7.5$ dB for motion deblurring systems, and $12$ dB for light field systems).}

\item{We use the MMSE as a metric to characterize the performance of the impulse and CI systems. However, the MMSE under GMM prior can not be computed analytically. We show that for CI systems, an analytic lower bound on MMSE (derived in \cite{flam}) can very closely approximate the exact MMSE and can used for analysis, see Figure \ref{fig:exactMMSEvsBoundsEDOF}.}

\end{tight_enumerate}

\subsection{Scope and Limitations}
\noindent \textbf{Image Formation Model.} Our analysis assumes a linear image formation model. 
Non-linear imaging systems, such as a
two/three photon microscopes and coherent imaging systems
are outside the scope of this paper. Nevertheless, our analysis covers a very large array of existing imaging systems 
\cite{raskar2006coded,MaskPaper,levin2007image,wagadarikar2008single,ihrke2010theory,ratner2007illumination}. 
We use a geometric optics model and ignore the effect of diffraction due to small apertures. 

\vspace{.05in}
\noindent\textbf{Noise Model.} We use an affine noise model to describe the combined effects of \textit{signal-independent} 
and \textit{signal-dependent} noise. Signal-dependent Poisson noise is approximated using a Gaussian noise model (as described in Section \ref{sec:Noise}).

\vspace{.05in}
\noindent\textbf{Single Image Capture.} We perform analysis of only single image CI 
techniques. Our results are therefore not applicable to multi-image capture 
techniques such as Hasinoff et al. \cite{Hasinoff:09} (EDOF), and Zhang et 
al. \cite{zhang2010defocusdenoise} (Motion Deblurring).

\vspace{.05in}
\noindent \textbf{Patch Based Prior.} Learning a GMM prior on entire images would require an impossibly 
large training set. To combat this problem, we train our GMM on image patches, and solve the image estimation
problem in a patch-wise manner. As a result, our technique requires that multiplexed measurements 
are restricted to linear combinations of pixels in a neighborhood smaller than the GMM patch size.

\vspace{.05in}
\noindent \textbf{Shift-Invariant Blur.} We analyze motion and defocus deblurring cameras under the assumption of a single known shift-invariant blur kernel. This amounts to the assumption that either the depth/motion is position-independent, or the blur is independent of depth/motion. We do not analyze errors due to inaccurate kernel estimation (for coded aperture and flutter shutter \cite{raskar2006coded,MaskPaper,levin2007image}) or due to the degree of depth/motion invariance (for focal sweep, cubic phase plate, motion invariant photography \cite{diffusionCoding,Baek:10,levin2008motion,cho2010motion}). 


%% file: relatedWork.tex
\textbf{Theoretical Analysis of CI systems:} Harwit and Sloan
\cite{harwit1979hadamard} analyzed coded imaging systems and have shown that, in
absence of photon noise, Hadamard and S-matrices are optimal. Wuttig and Ratner
et al. \cite{wuttig, ratner2007illumination, ratner2007optimal} then extended
the analysis to include both photon and read noise and showed that there is
significant gain in multiplexing only when the read noise dominates over photon
noise. Ihrke et. al. \cite{ihrke2010theory} analyzed the performance of
different light field cameras and color filter arrays. Tendero
\cite{TenderoThesis} has analyzed the performance of flutter shutter cameras
with respect of impulse imaging (short exposure imaging). Agrawal and Raskar
compared the performance of flutter shutter and motion invariant cameras
\cite{Agrawal:09}. Recently, Cossairt et. al. \cite{cossairt2013does,CAVE_0316}
has obtained optics independent upper bounds on performance for various CI
techniques. However, all the above works, do not analyze the
performance of CI systems when a signal prior is used for demultiplexing.
Cossairt et al. \cite{cossairt2013does} have performed empirical experiments to
study the effect of priors, but conclusions drawn based from simulations are
usually limited. 

\vspace{.05in} \noindent \textbf{Performance Analysis using Image Priors:}
 Zhou et al. \cite{Zhou:09} used a Gaussian
signal prior and Gaussian noise model to search for good aperture codes for defocus deblurring. 
Levin et al. \cite{levinCameraTradeoff}
have proposed the use of a GMM light field prior for comparing across different
light field (LF) camera designs. They used the mean square error as a metric for
comparing cameras. However, they do not take into account the effect of
signal dependent noise. 

Our approach is inspired in part by the recent analysis on the fundamental limits
of image denoising
\cite{chatterjee2010denoising,levin2011natural,levin12OptDenoise}, the only
papers we are aware of that directly address the issue of performance
bounds in the presence of image priors. Both of these recent results model
image statistics through a patch based image prior and derive lower bounds on
the MMSE for image denoising. We loosely follow the approach here
and extend the analysis to general computational imaging systems. In order to
render both computational tractability and generality, we use a GMM with a
sufficient number of mixtures to model the prior distribution on image patches.
Similar to \cite{chatterjee2010denoising,levin2011natural,levin12OptDenoise}, we
then derive bounds and estimates for the MMSE and use these to analyze CI
systems.

\vspace{.05in} \noindent\textbf{Practical Implications for CI systems:} Cossairt et. al.
\cite{cossairt2013does} analyzed CI systems based on
application (e.g. defocus deblurring or motion deblurring), lighting condition
(e.g. moonlit night or sunny day), scene properties (e.g. albedo, object
velocity) and sensor characteristics (size of pixels). They have shown that, for
commercial grade image sensors, CI techniques only improve performance significantly when the illumination is less than $125$ lux
(typical living room lighting). We extend these results to include the analysis
of CI systems with signal priors taken into account. Hasinoff et al.
\cite{Hasinoff:09} (in the context of EDOF) and Zhang et al.
\cite{zhang2010defocusdenoise} (in the context of motion deblurring)
analyzed the trade-off between denoising and deblurring for multi-shot imaging
within a time budget. We analyzed the trade-off between denoising and deblurring
for single shot capture.

%% file: ProbDef.tex
We consider linear multiplexed imaging systems that can be represented as 
\begin{equation}
y=Hx+n,
\label{eqn:linSys}
\end{equation}
where $y \in R^N$ is the measurement vector, $x \in R^N$ is the unknown signal we want to capture, $H$  is the $N \times N$  multiplexing matrix and $n$ is the observation noise.

\Subsection{Multiplexing Matrix $H$}
A large array of existing imaging systems follow a linear image formation model, such as flutter shutter \cite{raskar2006coded}, coded aperture \cite{MaskPaper,levin2007image,wagadarikar2008single}, plenoptic multiplexing \cite{ihrke2010theory}, illumination multiplexing \cite{schechner2007multiplexing}, and many others. The results of this paper can be used to analyze all such systems. 
In this paper, we analyze motion and defocus deblurring systems and multiplexed light field systems. For motion and defocus blur, we concentrate mostly on sytems that produce shift-invariant blur. For the case of 1D motion blur, the vectors $x$ and $y$ represent a scan line in a sharp and blurred image patch, respectively. The multiplexing matrix $H$ is a Toeplitz matrix where the rows contain the system point spread function. For the case of 2D defocus blur, the vectors $x$ and $y$ represent lexicographically reordered image patches, and the multiplexing matrix $H$ is block Toeplitz.  For the case of light field systems, $x$ and $y$ represent lexicographically reordered light field and captured 2D multiplexed image patch and $H$ matrix is a block Toeplitz matrix. 

\Subsection{Noise Model}
\label{sec:Noise}
To enable tractable analysis, we use an affine noise model \cite{schechner2007multiplexing,Hasinoff:09}. We model signal independent noise as a Gaussian random variable with variance $\sigma_{r}^2$. Signal dependent photon noise is Poisson distributed with parameter $\sigma_p^2$ equal to the average signal intensity at a pixel. We approximate photon noise by a Gaussian distribution with variance $\sigma_p^2$. This is a good approximation when $\sigma_p^2$ is greater than $10$. We also drop the pixel-wise dependence of photon noise and instead assume that the noise variance at every pixel is equal to the average signal intensity. For a given lighting and scene, if $J$ is the average pixel value in the impulse camera, then the photon noise variance is given by $\sigma_p^2=J$. For the same lighting and scene, the average pixel value for a CI camera (specified by multiplexing matrix $H$) is given by $C(H)J$, where $C(H)$ is the matrix light throughput, defined as the average row sum of $H$. Thus, the average photon noise variance for the CI system is $\sigma_p^2=C(H)J$ and the overall noise model is given by:

\begin{equation}
f(n)=\mathcal{N}(0,C_{nn}), \textrm{ } C_{nn}=(\sigma_r^2+C(H)J)\mathcal{I},
\label{eqn:noiseModel}
\end{equation}
where $\mathcal{I}$ is the identity matrix with dimension equal to the number of observed pixels. 

\Subsection{Signal Prior Model}
\label{sec:Prior}
In this paper, we choose to model scene priors using a GMM because of three characteristics: 

\begin{tight_itemize}
\item{\textbf{State of the art performance:}
GMM priors have provided state of the art results in various imaging  applications such as image denoising, deblurring and superresolution \cite{sapiro,colon}, still-image compressive sensing \cite{sapiro,csManifold}, light field denoising and  superresolution \cite{mitra2012light} and video compressive sensing \cite{carinVideoCS}. GMM is also closely related to the union-of-subspace model \cite{eldar2009,eldar2010} as each Gaussian mixture covariance matrix defines a principle subspace.}

\vspace{.1in}
\item{\textbf{Universal Approximation Property:}
GMM satisfies the universal approximation property i.e., (almost) any prior can be approximated by learning a GMM with a large enough number of mixture components \cite{GMMapproxAllPrior1,GMMapproxAllPrior2}. 
To state this concisely, consider a family of zero mean Gaussian distributions $\mathcal{N}_\lambda (x)$  with variance  $\lambda$. Let $p(x)$ be a prior probability density function with a finite number of discontinuities, that we want to approximate using a GMM distribution. Then the following Lemma holds:
\newtheorem{lemmaGMM}{Lemma}[section]
\begin{lemmaGMM}\label{lemmaGMM}
The sequence $p_\lambda (x)$ which is formed by the convolution of $N_\lambda (x)$ and $p(x)$
\begin{equation}
p_\lambda (x)=\int_{-\infty}^{\infty}\mathcal{N}_\lambda (x-u) p(u)d\textrm(u)
\end{equation}
converges uniformly to $p(x)$ on every interior sub-interval of $(-\infty,\infty)$.
\end{lemmaGMM}
This Lemma is a restatement of Theorem 2.1 in \cite{GMMapproxAllPrior1}. 
The implication of this Lemma is that priors for images, videos, light-fields and other visual signals can all be approximated using a GMM prior with appropriate number of mixture components, thereby allowing our framework to be applied to analyze a wide range of computational imaging systems.}

\vspace{.1in}
\item{\textbf{Analytical Tractability:} Unlike other state-of-the-art signal priors such as dictionary learning \cite{aharon2006rm, mairal} and BM3D \cite{dabov2007image}, we can analytically compute a good lower bound on MMSE \cite{flam} as described in section \ref{sec:exactVsBounds}. 
}
\end{tight_itemize}

\Subsection{Performance Characterization}
We characterize the performance of multiplexed imaging systems under (a) the noise model described in section \ref{sec:Noise} and (b) the scene prior model described in section \ref{sec:Prior}. For a given multiplexing matrix $H$, we will study two metrics of interest: (1) $mmse(H)$, which is the minimum mean squared error (MMSE) and (2) multiplexing SNR gain $G(H)$ defined as the SNR gain (in dB) of the multiplexed system $H$ over that of the impulse imaging system whose $H$-matrix is the identity matrix $I$:
\begin{equation}
G(H)=10log_{10}(\frac{mmse(I)}{mmse(H)}).
\label{eq:SNRgain}
\end{equation}

%% file: GMMPrior.tex

Mean Squared Error (MSE) is a common metric for characterizing the performance of linear systems under Bayesian setting. Among all estimators the MMSE estimator achieves the minimal MSE and we use the corresponding error (MMSE) for characterizing the performance of the CI systems. As discussed earlier in Section \ref{sec:ProbDef}, we model the signal using GMM prior and the noise using Gaussian distribution and compute the MMSE of the CI systems. Recently, Flam et al. \cite{flam} have derived the MMSE estimator and the corresponding error (MMSE) for linear systems under GMM signal prior and GMM noise model. Ours is thus a special case with the noise being Gaussian distributed, see section \ref{sec:Noise}. We present the expressions of the MMSE estimator and the corresponding error here, for derivations see \cite{flam}.


GMM distribution is specified by the number of Gaussian mixture components $K$, the probability of each mixture component $p_k$, and the mean and covariance matrix $(u_x^{(k)},C_{xx}^{(k)})$ of each Gaussian:
\begin{equation}
f(x)=\sum_{k=1}^K  p_k \mathcal{N}(u_x^{(k)},C_{xx}^{(k)}).
\end{equation} 
As discussed in section \ref{sec:Noise}, we model the signal independent and dependent noise as a Gaussian distributed $\mathcal{N}(0,C_{nn})$, see Eqn. \eqref{eqn:noiseModel}. From Eqn. \eqref{eqn:linSys}, the likelihood distribution of the measurement $y$ is given by $f(y|x)=\mathcal{N}(Hx,C_{nn})$. After applying Bayes rule, the posterior distribution $f(x|y)$ \textit{is also a GMM distribution} with new weights $\alpha^{(k)}(y)$ and new Gaussian distributions $f^{(k)}(x|y)$:

\begin{equation}
f(x|y)=\sum_{k=1}^K \alpha^{(k)}(y) f^{(k)}(x|y),
\label{eq:GMMposterior}
\end{equation}
where $f^{(k)}(x|y)$ is the posterior distribution of the $k^{th}$ Gaussian 

\begin{equation}
f^{(k)}(x|y)=\mathcal{N}(u_{x|y}^{(k)}(y),C_{x|y}^{(k)})
\label{eq:GMMGaussians}
\end{equation}
with mean 

\begin{equation}
u_{x|y}^{(k)}(y)= u_x^{(k)} + C_{xx}^{(k)} H^T (H C_{xx}^{(k)} H^T+C_{nn})^{-1} (y-Hu_x^{(k)}), 
\end{equation} 
and covariance matrix

\begin{equation}
C_{x|y}^{(k)}=C_{xx}^{(k)} - C_{xx}^{(k)} H^T(HC_{xx}^{(k)} H^T+C_{nn})^{-1}HC_{xx}^{(k)}.
\label{eqn:PosteriorCovariance}
\end{equation} 

The new weights $\alpha^{(k)}(y)$ are the old weights $p_k$ modified by the probability of $y$ belonging to the $k^{th}$ Gaussian mixture component

\begin{equation}
\alpha^{(k)}(y)=\frac{p_kf^{(k)}(y)}{\sum_{i=1}^K {p_if^{(i)}(y)}}, 
\label{eq:posteriorWt}
\end{equation} 
where $f^{(k)}(y)$, which is the probability of $y$ belonging to the $k^{th}$ Gaussian component, is given by:

\begin{equation}
f^{(k)}(y)=\mathcal{N}(y;Hu_x^{(k)},HC_{xx}^{(k)}H^T+C_{nn})
\label{eq:marginalObsPerMixture}
\end{equation}

The MMSE estimator $\hat{x}(y)$ is the mean of the posterior distribution $f(x|y)$, i.e.,
\begin{equation}
\hat{x}(y)=\sum_{k=1}^K \alpha^{(k)}(y) u_{x|y}^{(k)}(y).
\label{eqn:GMM_MMSE}
\end{equation}

The corresponding MMSE is given by
\begin{equation}
mmse(H) =  E||x-\hat{x}(y)||^2\\ 
\label{eq:GMMMSEDef}
\end{equation}
As shown in \cite{flam} (see Eqns. 26-29 in \cite{flam}), the $mmse(H)$ can be written as a sum of two terms: an \emph{intra-component} error term and an \emph{inter-component} error term. 
\begin{equation}
\begin{split}
mmse(H)  =  \sum_{k=1}^K p_k Tr(C_{x|y}^{(k)}) \\ 
 + \sum_{k=1}^K p_k \int_y||\hat{x}(y)-u_{x|y}^{(k)}(y)||^2 f^{(k)}(y)\textrm{d}y,
\end{split}
\label{eqn:GMM_MSE}
\end{equation}
where $Tr$ denotes matrix trace. Any given observation $y$ is sampled from one of the $K$ Gaussian mixture components. The first term in Eqn. \eqref{eqn:GMM_MSE} is the intra-component error, which is the MSE for the case when $y$ has been correctly identified with its original mixture component. The second term is the inter-component error, which is the MSE due to inter-component confusion. The proof for the above decomposition is given in \cite{flam}. Note that the first term in Eqn. \eqref{eqn:GMM_MSE} is independent of the observation $y$ and depends only on the multiplexing matrix $H$, the noise covariance $C_{nn}$, and the learned GMM prior parameters $p_k$, and $C_{xx}$ and can be computed analytically. However, we need Monte-Carlo simulations to compute the second term in Eqn. \eqref{eqn:GMM_MSE}. 

%
%

%% file: frmAnalCIsystems.tex

We study the performance of various CI systems under the practical consideration of illumination conditions and sensor characteristics.

\Subsection{Performance Characterization} Computational Imaging (CI) systems
improve upon traditional imaging systems by allowing more light to be captured
by the sensor. However, captured images then require decoding, which typically
results in noise amplification. To improve upon performance, the benefit of
increased light throughput needs to outweigh the degradations caused by the
decoding process. The combined effect of these two processes is measured as the
SNR gain.
Following the approach of \cite{cossairt2013does}, we measure SNR gain relative to impulse imaging. 
%
However, the analysis in \cite{cossairt2013does} does not address the fact that
impulse imaging performance can be significantly improved upon by state of the
art image denoising methods
\cite{dabov2007image,chatterjee2010denoising,levin2011natural}. We correct this
by denoising our impulse images using the GMM prior. The effect this has on
performance is clearly seen in Figure \ref{fig:motiveFig}. The dotted blue line
corresponds to impulse imaging without denoising, while the solid blue line
corresponds to impulse imaging after denoising using the GMM prior. Thus, the
results presented in Figures \ref{fig:compareEDOFsystems},
\ref{fig:compareMotDeblurSystems} show the performance improvements obtained due
to CI over that of impulse imaging with state of the art
denoising. Another important result of this paper, is that much like
\cite{chatterjee2010denoising,levin2011natural}, we are also able to quantify
the significant performance improvements that can be obtained through image
denoising.

\Subsection{Scene Illumination Level}
The primary variable that controls the SNR of impulse imaging is the scene illumination level.
As discussed in section \ref{sec:Noise}, we consider two noise types: photon noise (signal dependent) and read noise (signal independent). Photon noise is directly proportional to the scene illumination level, whereas, read noise is independent of it. 
At low illumination levels, read noise dominates the photon noise but, since signal power is low, the SNR is typically low.
At high scene illumination levels, photon noise dominates the read noise. 
Recognizing this, we compare CI techniques to impulse imaging over a wide range of scene illumination levels.

\Subsection{Imaging System Specification} \label{sec:camSpecs} 
Given the scene illumination level $I_{src}$ (in lux), the average scene reflectivity ($R$) and the camera parameters such as the f-number ($F/\#$), exposure time ($t$), sensor quantum efficiency ($q$), and pixel size ($\delta$),  the average signal level in photo-electrons ($J$)  of the impulse camera is given by \cite{cossairt2013does}\footnote{The signal level will be larger for CI techniques. The increase in signal is encoded in the multiplexing matrix $H$, as discussed in Section \ref{sec:GMMPrior}}:
\begin{equation}
J=10^{15}(F/\#)^{-2}tI_{src}Rq(\delta)^2.
\label{eqn:illumToPhotons}
\end{equation}
In our experiments, we assume an average scene reflectivity of $R=0.5$ and sensor quantum efficiency of $q=0.5$, aperture setting of $F/11$ and exposure time of $t=6$ milliseconds, which are typical settings in consumer photography.  

Sensor characteristics impact the SNR directly: sensors with larger pixels
produce a higher SNR at the same scene illumination level. 
Here, we choose three different example
cameras that span the a wide range of consumer imaging devices: $1)$ a high end
SLR camera, $2)$ a machine vision camera (MVC) and $3)$ a smartphone camera
(SPC). For each of these example camera types, we choose parameters that are
typical in the marketplace today: sensor pixel size: $\delta_{SLR}=8 \mu m$ for
the SLR camera, $\delta_{MVC}=2.5 \mu m$ for the MVC, and $\delta_{SPC}=1\mu m$
for the SPC. We also assume a sensor read noise of $\sigma_r=4e^-$ which is
typical for today's CMOS sensors.
The x-axis of the
plots shown in Figures \ref{fig:motiveFig}, \ref{fig:compareEDOFsystems} and
\ref{fig:compareMotDeblurSystems} for SLR, MVC and the SPC are simply shifted
relative to one another.

\Subsection{Experimental Details}

The details of the experimental setup are as follows

\begin{tight_itemize}
\item{\textbf{Learning:} We learn GMM patch priors from a large collection of training patches. For EDOF and motion deblurring experiments, we learn the prior model using the $200$ training images from Berkeley segmentation dataset \cite{bsd}. For LF experiment, we use the Standford light field dataset \footnote{http://lightfield.stanford.edu/lfs.html} for learning the prior model. For learning we use a variant of the Expectation Maximization approach to ascertain the model parameters. We also test that the learned model is an adequate approximation of the real image prior by performing rigorous statistical analysis and comparing performance of the learned prior with state of the art image denoising methods \cite{dabov2007image}. Since, we learn GMM prior on image patches, patch size is an important parameter that needs to be chosen carefully. We choose patch size based on two considerations: $1)$ patch size should be bigger than the size of local multiplexing (blur kernel size) and $2)$ it is difficult to learn good prior for large patch sizes. In the analysis of EDOF systems, we have chosen blur kernel of  $11 \times 11$ for focal sweep \cite{Nagahara:08}, coded aperture by Zhou et al. \cite{Zhou:09} and coded aperture by Levin et al. \cite{levin2007image}. We experimented with different patch sizes ($>11\times 11$) and found that patch size of $24 \times 24$ gives the best simulation results. Thus, for our experiments on EDOF systems, we chose the patch size to be $24\times 24$. In the analysis of motion deblurring systems, we have chosen the flutter shutter kernel size to be $1\times 33$ and motion invariant kernel size to be $1\times 9$. After experimenting with different GMM patch sizes ($> 1\times 33$), we found that patch size of $4\times 256$ gives the best simulation results and hence we chose that patch size for analysis of motion deblurring systems. For LF experiment we use GMM patch prior of size $16\times 16 \times 5 \times 5$  as proposed in \cite{mitra2012light}. Further in-depth study is required for optimal choice of patch sizes for each application. However, this is outside the scope of this paper and is a topic of focus for future study.} 
\vspace{.1in} 
\item{\textbf{Analytic Performance metric:}  Analytic performance is compared using the MMSE metric. Once the MMSE is computed for the impulse and CI systems, we compute the multiplexing SNR gain in dB using Eqn. \eqref{eq:SNRgain}. The analytic multiplexing gain for various CI systems are shown in Figures \ref{fig:motiveFig}, \ref{fig:compareEDOFsystems} and \ref{fig:compareMotDeblurSystems}(a).}
\vspace{.1in}
\item{\textbf{Analytic Performance without Prior:} To calculate the performance of CI systems without signal priors taken into account, we compute the MSE as:
 \begin{equation} 
mse(H)=Tr(H^{-1} C_{nn} H^{-T}),
\end{equation}	
where $H$ is the corresponding multiplexing matrix and $C_{nn}$ is the noise covariance matrix.}
\vspace{.1in}
\item{\textbf{Analytic Performance with Prior:} The analytic performance of CI systems with priors taken into account is computed as described in Section \ref{sec:GMMPrior} (Eqn. \eqref{eqn:GMM_MSE}). These results are shown in Figures \ref{fig:motiveFig}, \ref{fig:compareEDOFsystems} and \ref{fig:compareMotDeblurSystems}(a).}
\vspace{.1in}
\item{\textbf{Simulations Results for Comparison:} In order to validate our analytic predictions, we also performed extensive simulations. In our simulations, we used the MMSE estimator,  Eqn. \eqref{eqn:GMM_MMSE}, to reconstruct the original (sharp) images. The MMSE estimator has been shown to provide state of art results for image denoising \cite{levin2011natural}, and here we extend these powerful methods for general demultiplexing. For comparison we also perform simulations using BM3D \cite{dabov2007image}. For EDOF and motion deblurring simulations we use the image deblurring version of the BM3D algorithm \cite{Dabov08deblur}, and for light field simulations we first perform linear reconstruction and then denoise it using the BM3D algorithm. Some images of our simulation experiments are shown in Figures \ref{fig:compareEDOFsimuLowLight}, \ref{fig:compareMotDeblurSimuLowLight} and \ref{fig:compareLFSimuLowLight}, providing visual and qualitative comparison between CI and traditional imaging techniques. The simulation results are consistent with our analytic predictions and show that CI provides performance benefits over a wide range of imaging scenarios.}
\end{tight_itemize}

%% file: perfEDOFsystems.tex

\begin{figure}[tbh]
	\centering
		\includegraphics[width=\columnwidth]{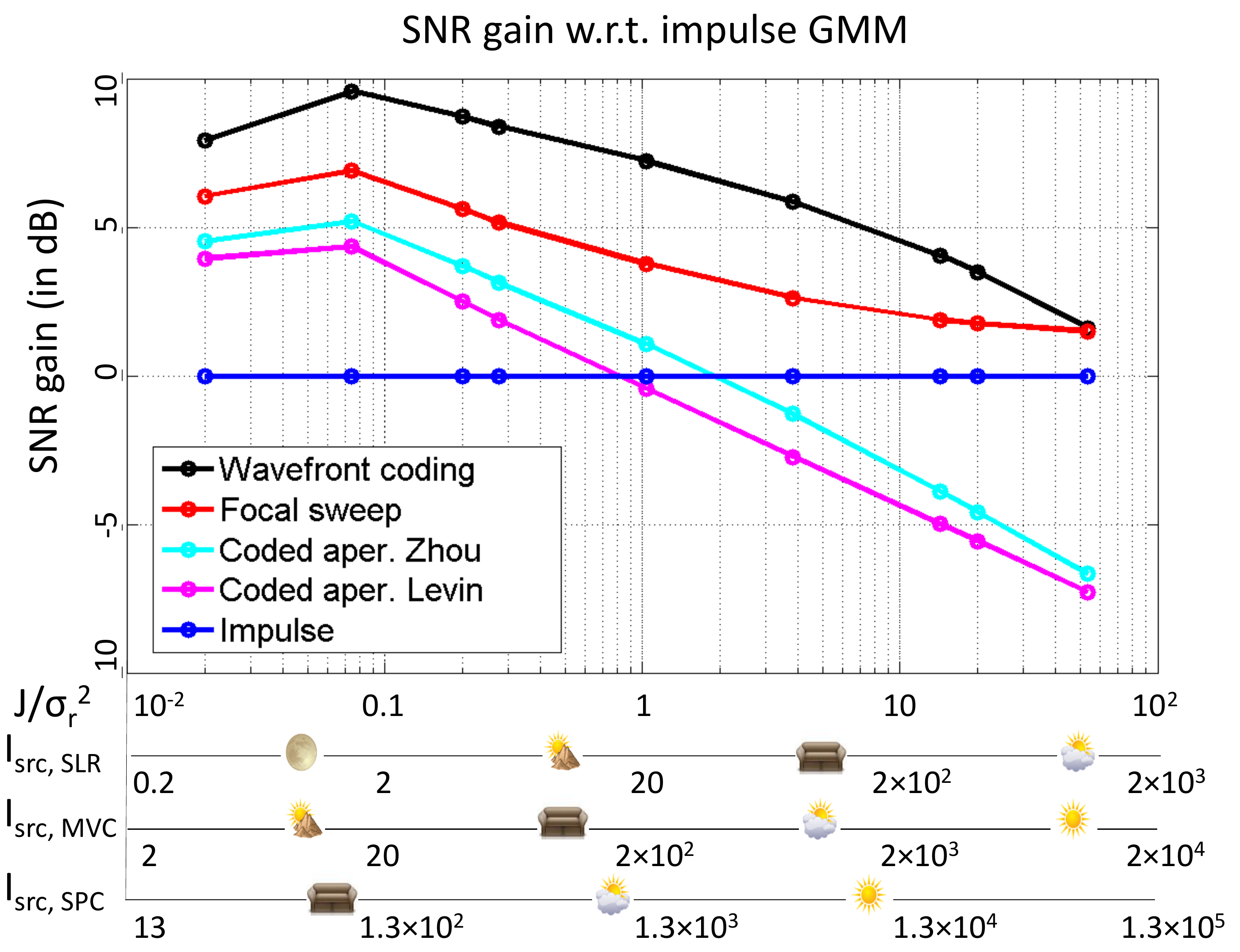}
\vspace{-.1in}
	\caption{\emph{Analytic performance of EDOF systems under signal prior:} We plot the SNR gain of various EDOF systems at different photon to read noise ratios ($J/\sigma_r^2$). In the extended x-axis, we also show the effective illumination levels (in lux) required to produce the given $J/\sigma_r^2$ for the three camera specifications: SLR, MVC and SPC. The EDOF systems that we consider are:  cubic phase wavefront coding \cite{dowski1995extended}, focal sweep camera \cite{Nagahara:08}, and the coded aperture designs by Zhou et al. \cite{Zhou:09} and Levin et al. \cite{levin2007image}. Signal priors are used to improve performance for both CI and impulse cameras. Wavefront coding gives the best performance amongst the compared EDOF systems and the SNR gain varies from a significant $9.6$ dB at low light conditions to $1.6$ dB at high light conditions. This demonstrates the benefits of multiplexing beyond the use of signal priors, especially at low light condtions. For corresponding simulations, see figure \ref{fig:compareEDOFsimuLowLight}.} 
\label{fig:compareEDOFsystems}
\vspace{-.1in}
\end{figure}

\begin{figure*}[tbh]
	\centering
		\includegraphics[width=2\columnwidth]{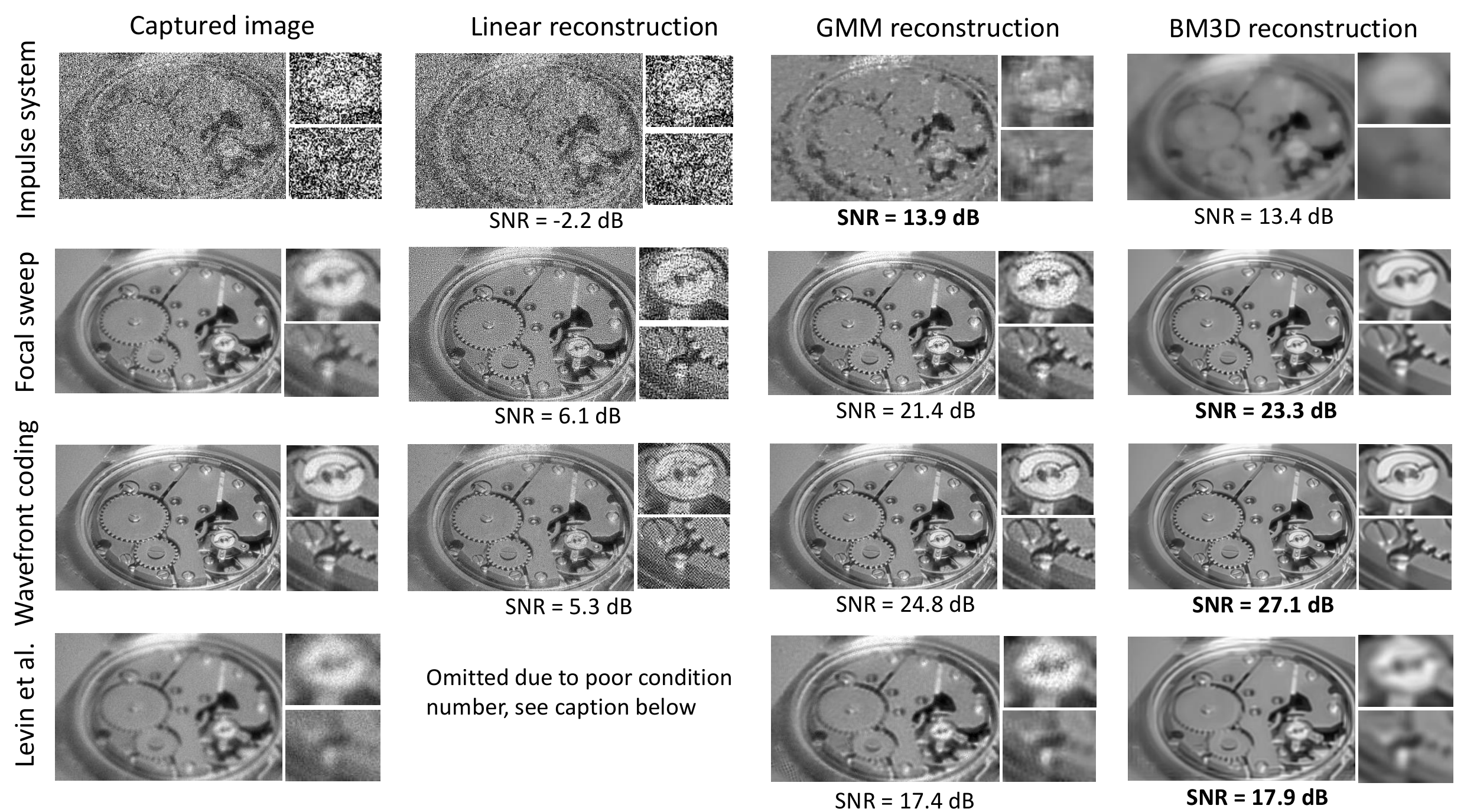}
\vspace{-.1in}
	\caption{\emph{Simulation results for EDOF systems at low light condition (photon to read noise ratio of $J/\sigma^2=0.2$):} We show reconstruction results for many EDOF systems with and without signal prior. We do not show linear reconstruction for coded aperture design of Levin et al. \cite{levin2007image}, because the corresponding $H$ matrix has poor condition number. The frequency spectrum of the designed code has zeroes, which are good for estimating depth from defocus but not good for reconstruction. Note that the use of signal prior significantly increases the performance of both impulse and CI systems. Using GMM prior, focal sweep \cite{Nagahara:08}, wavefront coding \cite{dowski1995extended} and coded aperture design of Levin et al.  \cite{levin2007image} produce SNR gains (w.r.t. impulse system) of $7.5$ dB, $10.9$ dB and $3.5$ dB respectively, which are significant gains. For impulse imaging and coded aperture of Levin, GMM and BM3D give similar reconstruction SNR, where as for the focal sweep and wavefront coding BM3D reconstructions are $~2$ dB better than the GMM reconstruction.}
\label{fig:compareEDOFsimuLowLight}
\vspace{-.2in}
\end{figure*}

We study the SNR gain of various EDOF systems with and without the use of signal priors. For the signal prior, we learn a  GMM patch prior of patch size $24\times 24$ with $1500$ Gaussian mixtures. First we study the performance of a particular EDOF system, focal sweep \cite{Nagahara:08}, and compare it with impulse imaging. We assume the aperture size of the focal sweep system to be $11\times11$ times bigger than that of the impulse camera, corresponding to an aperture setting of $F/1$. Hence, the light throughput of focal sweep is about $121$ times that of the impulse camera. Figure \ref{fig:motiveFig} shows the analytical SNR gain for focal sweep and impulse cameras with and without using signal prior. The plot shows performance measured relative to impulse imaging without a signal prior (no denoising). Without signal prior, focal sweep has a huge SNR gain over impulse imaging at low photon to read noise ratio, $J/\sigma_r^2$. This is consistent with the result obtained in \cite{cossairt2013does}. However, given that most state-of-the-art reconstruction algorithms are based on signal priors, these gains are unrealistic. When the signal prior is taken into account, we get realistic gains of $7$ dB at low light conditions. From the plot it is also clear that the the use of prior increases SNR much more than does multiplexing. 

Further, we study the performance of various other EDOF systems such as cubic phase wavefront coding \cite{dowski1995extended}, and the coded aperture designs by Zhou et al. \cite{Zhou:09} and Levin et al. \cite{levin2007image} \footnote{The performance of coded aperture systems reported here is overoptimistic because we assume perfect kernel estimation, as discussed in Section \ref{sec:Intro}.}. 
Figure \ref{fig:compareEDOFsystems} shows the SNR gain (in dB) of these EDOF systems with respect to impulse imaging \emph{under signal prior} (denoising). Amongst these systems, wavefront coding gives the best performance with SNR gain varying from a significant $9.6$ dB at low light conditions to $1.6$ dB at high light conditions. For corresponding simulations, see figure \ref{fig:compareEDOFsimuLowLight} \footnote{The watch image for simulation is obtained courtesy Ivo Ihrke and Matthias B. Hullin.}. Again from the simulations we can conclude that the use of signal prior significantly increases the performance of both impulse and CI systems and that wavefront coding gives significant performance gain over impulse imaging even after taking signal priors into account. In figure \ref{fig:compareEDOFsimuLowLight}, we also show reconstructions using BM3D \cite{dabov2007image}. For the impulse system and coded aperture of Levin, GMM and BM3D give similar reconstruction SNR, where as for the focal sweep and wavefront coding BM3D reconstructions are $~2$ dB better than the GMM reconstruction.


\noindent \textbf{Practical Implications:}
The main conclusions of our analysis are 
\begin{tight_itemize}
\item{\emph{The use of signal priors improves the performance of both CI and impulse imaging significantly.}}
\item{\emph{Wavefront coding gives the best performance amongst the compared EDOF systems and the SNR gain varies from a significant $9.6$ dB at low light conditions to $1.6$ dB at high light conditions. This demonstrates the benefits of multiplexing beyond the use of signal priors, especially at low light condtions.}}
\end{tight_itemize}

%% file: perfMotDeblurSystems.tex
\begin{figure*}[tbh!]
	\centering
	\includegraphics[width=2\columnwidth]{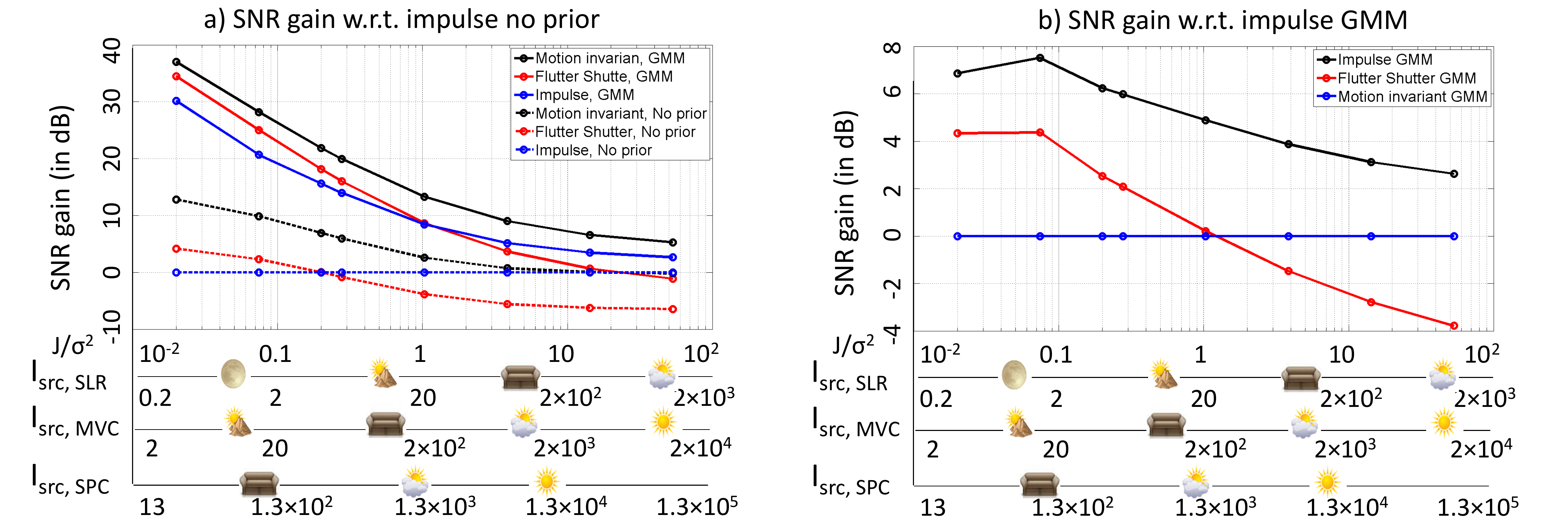}
\vspace{-.1in}
	\caption{\emph{Analytic performance of motion deblurring systems:} We study the performance of motion invariant \cite{levin2008motion}, flutter shutter \cite{raskar2006coded} and impulse cameras with and without the use of signal priors. From subplot (a), it is clear that SNR gain due to signal prior is much more than due to multiplexing. However, after taking into account the effect of signal prior, multiplexing still produce significant SNR gains as shown in subplot (b). Motion invariant imaging produces SNR gains ranging from $7.5$ dB at low light conditions to $2.5$ dB at high light conditions. For corresponding simulations, see figure \ref{fig:compareMotDeblurSimuLowLight}.}
\label{fig:compareMotDeblurSystems}
\vspace{-.1in}
\end{figure*}

\begin{figure*}[tbh]
	\centering
		\includegraphics[width=2\columnwidth]{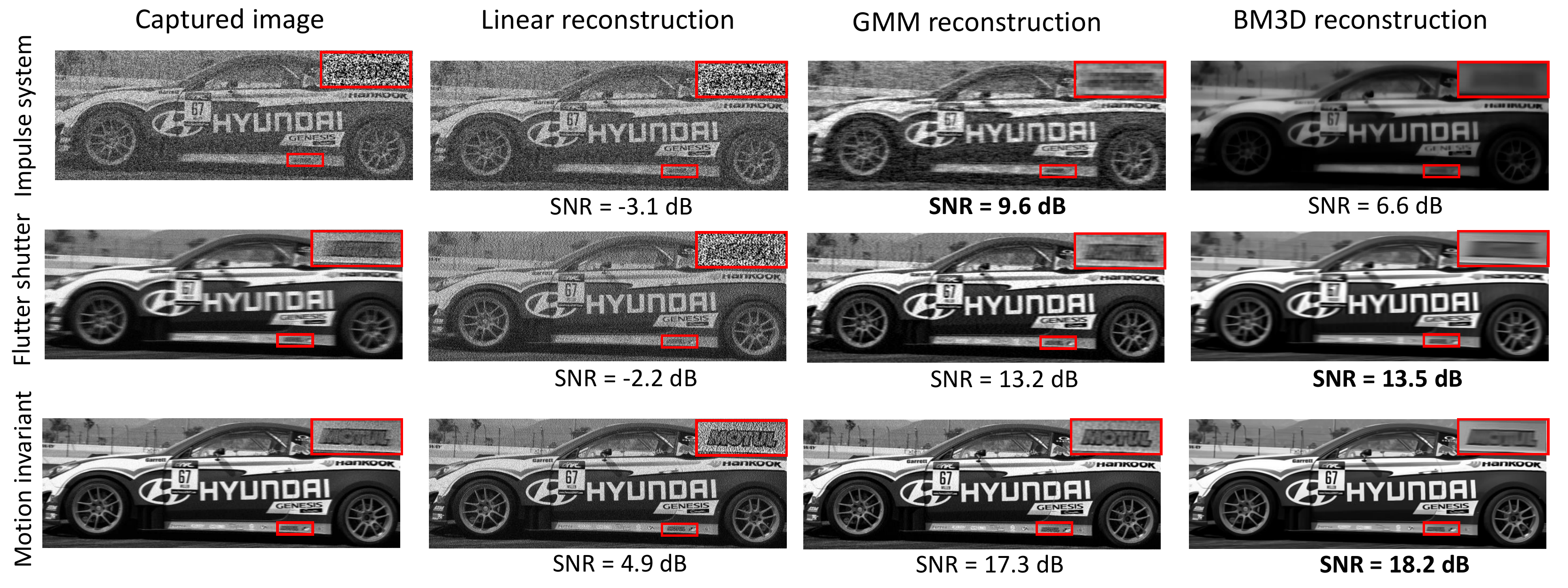}
\vspace{-.1in}
	\caption{\emph{Simulation results for motion deblurring systems at low light condition (photon to read noise ratio of $J/\sigma^2=0.2$).} Using GMM prior, flutter shutter \cite{raskar2006coded} and motion invariant system \cite{levin2008motion} produce SNR gains (w.r.t. impulse system) of $3.6$ dB and $7.7$ dB respectively. For the impulse system, GMM reconstruction is $3$ dB better than the BM3D reconstruction, where as for the flutter shutter and motion invariant system, GMM and BM3D produce similar results. }
\label{fig:compareMotDeblurSimuLowLight}
\vspace{-.2in}
\end{figure*}

We study the performance of two motion deblurring systems: the flutter shutter \cite{raskar2006coded} and motion invariant camera \cite{levin2008motion}. Again, we focus our attention on the case where signal priors are taken into account. For this experiment, we learn a GMM patch prior, of patch size $4\times256$, with $1500$ Gaussian mixtures. For the motion deblurring cameras, we set the exposure time to be $33$ times that of the impulse camera, corresponding to an exposure time of $200$ milliseconds. The binary flutter shutter code that we used in our experiment has $15$ 'ones' and hence the light throughput is $15$ times that of the impulse imaging system. The light throughput of the motion invariant camera is $33$ times that of the impulse camera. Figure \ref{fig:compareMotDeblurSystems}(a) shows the analytic SNR gain (in dB) of the motion deblurring systems with respect to impulse imaging without signal prior \footnote{Flutter shutter performance reported here is overoptimistic because we assume perfect kernel estimation, as discussed in Section \ref{sec:Intro}.}. Clearly, the SNR gain due to signal prior is much more than that due to multiplexing. However, after taking into account the effect of signal prior, multiplexing still produce significant SNR gain as shown in \ref{fig:compareMotDeblurSystems}(b). Motion invariant imaging produces SNR gains ranging from $7.5$ dB at low light conditions to $2.5$ dB at high light conditions. Figure \ref{fig:compareMotDeblurSimuLowLight} show the corresponding simulation results.  At the low photon to read noise ratio of $J/\sigma_r^2=0.2$, motion invariant imaging performs $7.7$ dB better than impulse imaging. We also show simulation results using BM3D reconstruction. For impulse system, GMM reconstruction is $3$ dB better than the BM3D reconstruction, where as for the CI systems, both the reconstructions are similar. 

\vspace{.1in}
\noindent \textbf{Practical Implications:}
The main conclusion of our analysis is 
\begin{tight_itemize}
\item{\emph{Motion invariant imaging produces SNR gains ranging from $7.5$ dB at low light conditions to $2.5$ dB at high light conditions.}}
\end{tight_itemize}

%% file: perfLF.tex
\begin{figure*}[tbh!]
	\centering
	\includegraphics[width=2\columnwidth]{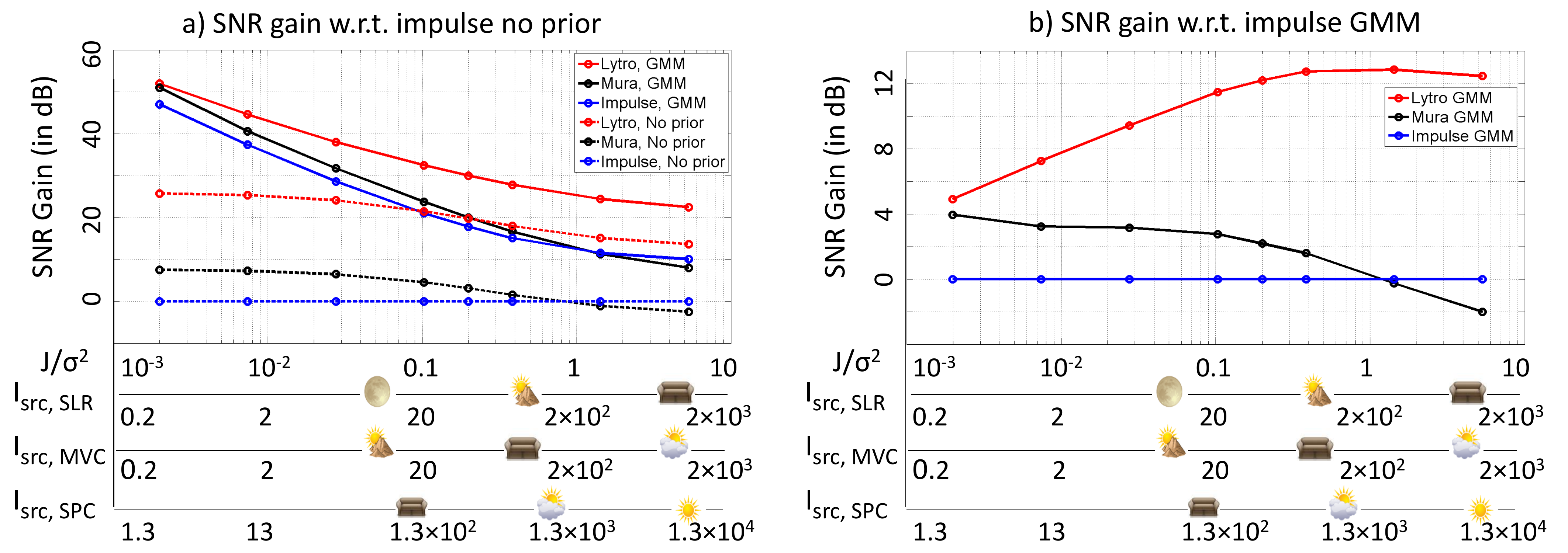}
\vspace{-.1in}
	\caption{\emph{Analytic performance of light field cameras:} We study the performance of the micro-lenses array based Lytro camera \cite{ng2005light} and MURA mask based light field camera \cite{lanman2008shield} against the light field impulse system (a pin-hole array mask camera). As in the case of EDOF and motion deblurring systems, the gain due to signal prior is much more than that due to multiplexing. Note that the SNR gain (w.r.t. impulse system) for Lytro is high even at high light levels. This is true for both with and without signal prior cases. This is because the multiplexing matrix of Lytro is a scaled version of the impulse multiplexing matrix, where the scale (light throughput) is given by the ratio of the lenselet area to the pinhole area. In our set-up, the light throughput of Lytro is about $20$ times that of the impulse system and so it is always about $13$ dB better than the impulse system. For corresponding simulations, see Figure \ref{fig:compareLFSimuLowLight}.}
\label{fig:analSnrGainLF}
\vspace{-.2in}
\end{figure*}

\begin{figure*}[tbh]
	\centering
		\includegraphics[width=2\columnwidth]{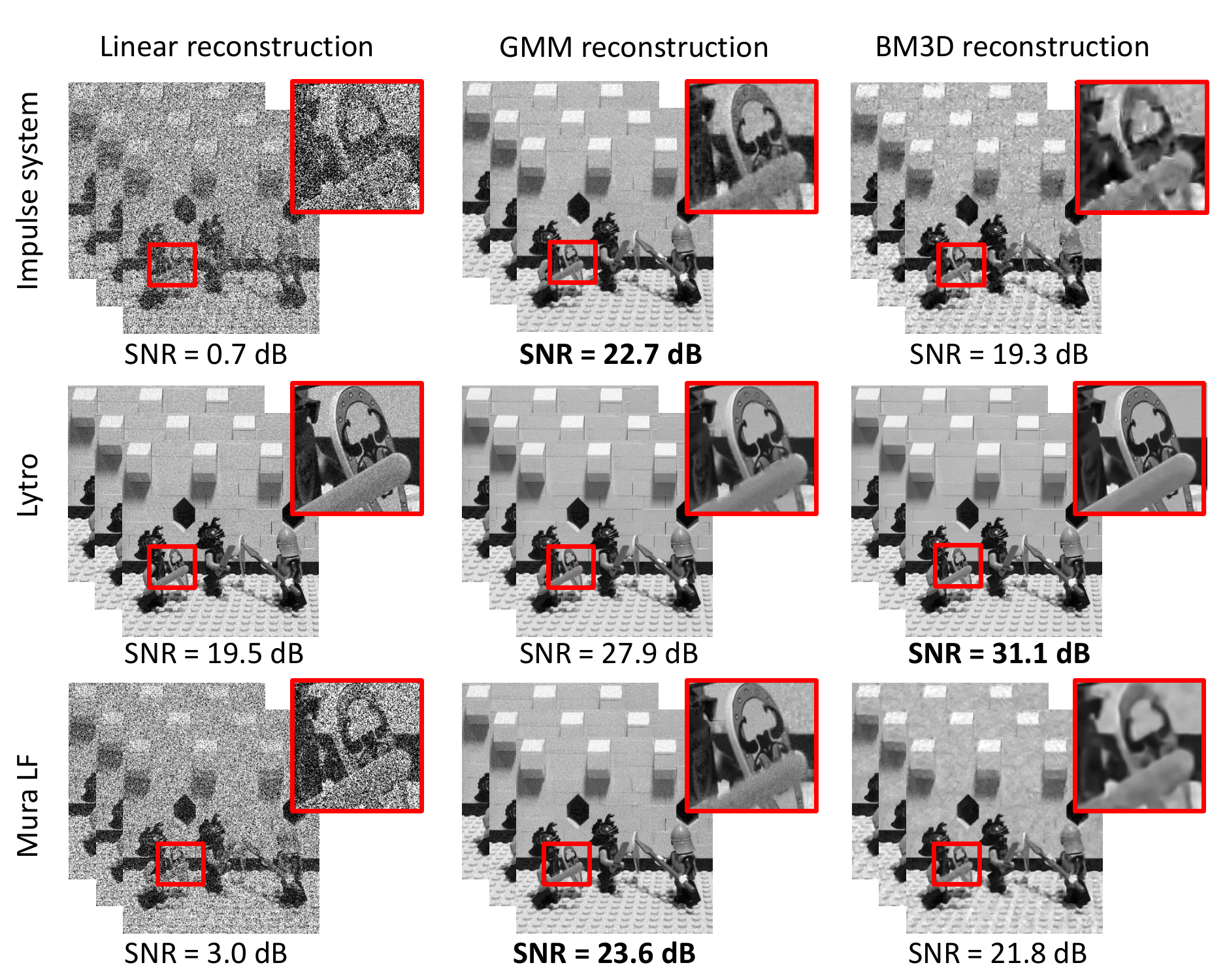}
\vspace{-.1in}
	\caption{\emph{Simulation results forlight field cameras at low light condition (photon to read noise ratio of $J/\sigma^2=0.2$).} Using GMM prior, Lytro \cite{ng2005light} and MURA mask based light field camera \cite{lanman2008shield} produce SNR gains (w.r.t. impulse systems) of $5.2$ dB and $0.9$ dB respectively.}
\label{fig:compareLFSimuLowLight}
\vspace{-.2in}
\end{figure*}

We study the performance of two light field cameras: $1)$ the micro-lenses array based Lytro camera \cite{ng2005light} and $2)$ MURA mask based light field camera \cite{lanman2008shield}. The corresponding impulse system is a pin-hole mask array placed at the micro lenses array location. We use the GMM patch prior of size $16 \times 16 \times 5 \times 5$ as proposed in \cite{mitra2012light}, which learns a Gaussian component for each disparity value between the LF views. In our experiment we chose $11$ disparity values ranging from $-5:5$ and thus we learn GMM with $11$ components. For the MURA mask based LF we use tiled MURA mask with the basic tile of size $5\times 5$. The multiplexing matrix $H$ corresponding to the pin-hole mask array LF is the identity matrix, whereas the multiplexing matrix of Lytro is a scaled version of identity with the scale (light throughput) given by the ratio of the lenselet area to the pinhole area. For MURA based LF,  the captured $2$-D multiplexed image is obtained by inner product between the $5\times 5$ angular dimension of the LF and cyclically shifted versions of the basic $5\times5$ MURA mask and the $H$ matrix is constructed keeping this structure in mind. The locality of MURA-based reconstruction that enables a patch approach was established in \cite{ihrke2010theory}. Figure \ref{fig:analSnrGainLF}(a-b) shows the analytic SNR gains for the two CI systems w.r.t. the impulse system. As in the case of EDOF and motion deblurring systems, the gain due to signal prior is more than that due to multiplexing. Note that the SNR gain of Lytro is high even at high light levels. This is true for both with and without signal prior cases. This is because the multiplexing matrix of Lytro is a scaled version of the impulse multiplexing matrix. In our set-up, the light throughput of Lytro is about $20$ times that of the impulse system and so it is always about $13$ dB better than the impulse system. Figure \ref{fig:compareLFSimuLowLight} show corresponding simulations.  From our analysis, we conclude that Lytro provides significant SNR gain at high light levels, but similar performance to MURA at low light levels. However, we should keep in mind that the systems analyzed here all trade-off the spatial resolution for angular resolution and hence capture low spatial resolution light field data. There are designs that captures full spatial resolution light field data \cite{marwahCompLF}, but we have not analyzed them as the scope of this paper is limited to analyzing fully-determined system for which we can define a corresponding impulse system.  

\vspace{.1in}
\noindent \textbf{Practical Implications:}
The main conclusion of our analysis is 
\begin{tight_itemize}
\item{\emph{Lytro provides significant SNR gain at high light levels, but similar performance to MURA at low light levels.}}
\end{tight_itemize}

%% file: exactMMSEvsBounds.tex
The exact expression for MMSE is given by Eqn. \eqref{eqn:GMM_MSE}. As discussed in section \ref{sec:GMMPrior}, the first term depends only on the multiplexing matrix $H$, the noise covariance $C_{nn}$, and the learned GMM prior parameters $p_k$ and $C_{xx}$ and can be computed analytically. But we need to perform Monte-Carlo simulations to compute the second term. However, we can use the analytic first term as a lower bound on MMSE (and hence upper bound on SNR), i.e.,

\begin{equation}
mmse(H) \ge \sum_{k=1}^K p_k Tr(C_{x|y}^{(k)}).
\label{eqn:lowerBoundGMM_MSE}
\end{equation} 

Flam et al. \cite{flam} have also provided an upper bound for the MMSE, see Theorem 1 in \cite{flam}. They have shown that the LMMSE (linear MMSE) estimation error is an upper bound of the MMSE error. The LMMSE estimation error $lmmse$ is given by:

\begin{equation}
lmmse(H)=Tr(C_{xx} - C_{xx} H^T(HC_{xx} H^T+C_{nn})^{-1}HC_{xx}),
\label{eq:upperBoundMMSE}
\end{equation}
where

\begin{align*}
C_{xx} &=\sum_{k=1}^K p_k (C_{xx}^{(k)} + u_x^{(k)}u_x^{(k)}) -u_x u_x^T \\
u_x &= \sum_{k=1}^K p_k u_x^{(k)}.
\end{align*}

\begin{figure*}[tbh]
	\centering
	\includegraphics[width=2\columnwidth]{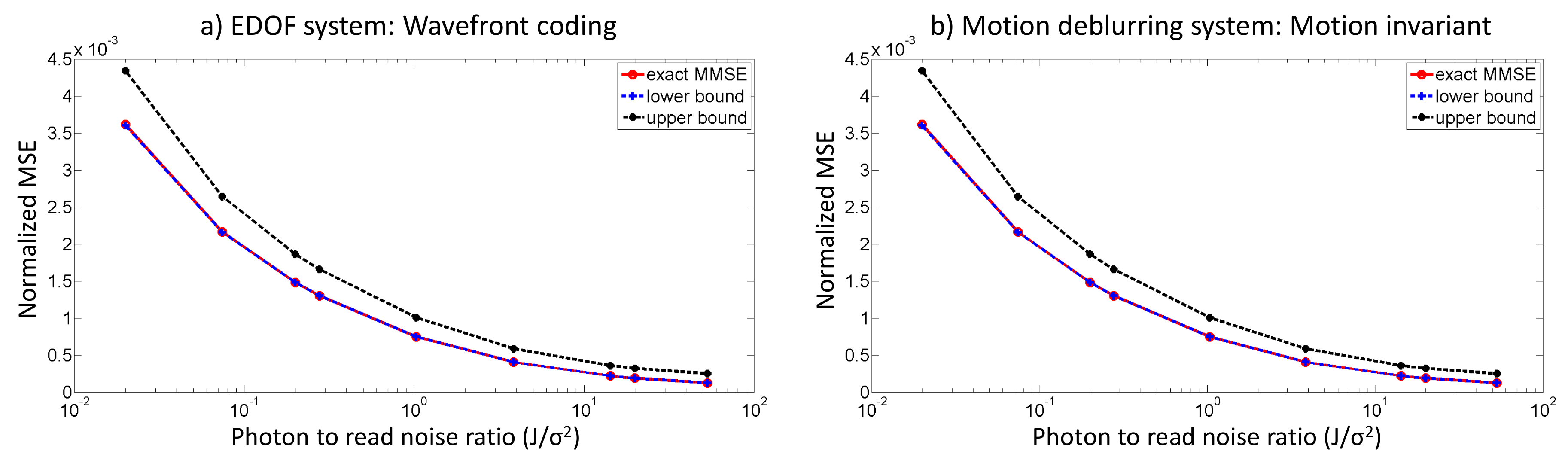}
	\caption{ \emph{Exact MMSE vs. its lower and upper bounds:} We compare the exact MMSE with it's analytic lower bound given by Eqn. \eqref{eqn:lowerBoundGMM_MSE} and upper bound given by Eqn. \eqref{eq:upperBoundMMSE}. Subplot (a) shows that, for the wavefront coding system, the lower bound is a very good approximation for the MMSE over the range $0.01<=J/\sigma^2<=100$. We do not show the corresponding plots for other EDOF systems as they are very similar to the wavefront coding system. Subplot (b) shows that the same conclusion holds for motion invariant system (and flutter shutter, not shown here). Thus, we can use the analytic lower bound for MMSE for both analysis and design of CI systems over a wide range of lighting levels.}
\label{fig:exactMMSEvsBoundsEDOF}
\end{figure*}

We compare the exact MMSE with its analytic lower and upper bounds given by Eqn. \eqref{eqn:lowerBoundGMM_MSE} and Eqn. \eqref{eq:upperBoundMMSE}, respectively, for the EDOF and motion deblurring systems. Figure \ref{fig:exactMMSEvsBoundsEDOF}(a) shows that, for the wavefront coding system, the lower bound is a very good approximation for the MMSE over the range $0.01<=J/\sigma^2<=100$. We do not show the corresponding plots for other EDOF systems as they are very similar to the wavefront coding system. Figure \ref{fig:exactMMSEvsBoundsEDOF}(b) shows that the same conclusion holds for motion invariant system. Note that though for these systems the lower bound is a very good approximation of the MMSE over the shown illumination range, it does not mean that this holds true for all light levels or for all systems. There are three factors which determine how well the lower bound can approximate the MMSE: 1) the multiplexing system $H$ (fully determined $H$ matrices are less likely to produce inter-component error as compared to under-determined systems), 2) noise level (large noise will lead to more inter-component error) and 3) location of Gaussian components in the GMM prior model. 

From the above experiment, we conclude that we can use the analytic lower bound expression for computing the MMSE of many EDOF and motion deblurring systems over a wide range of lighting condition. This also suggests that we can use the analytic lower bound expression of MMSE, Eqn. \eqref{eqn:lowerBoundGMM_MSE}, for solving the optimal CI design problem, i.e., finding the $H$ that minimizes Eqn. \eqref{eqn:lowerBoundGMM_MSE}, over a wide range of light levels.

%% file: Discussions.tex
We present a framework to comprehensively analyze the performance of CI systems. Our framework takes into account the effect of multiplexing, affine noise and signal priors. We model signal priors using a GMM, which can approximate almost all prior signal distributions. More importantly, the prior is analytically tractable. We use the MMSE metric to characterize the performance of any given linear CI system. Our analysis allows us to determine the increase in performance of CI systems \textit{when signal priors are taken into account}. We use our framework to analyze several CI techniques, specifically, EDOF, motion deblurring and light field cameras. Our analysis reveals that: $1)$ Signal priors increase SNR more than multiplexing, and $2)$ Multiplexing gain (above and beyond that due to signal prior) is significant especially at low light conditions. Moreover, we use our framework to establish the following practical implications: $1)$ Amongst the EDOF systems analyzed in the paper, Wavefront coding gives the best performance with SNR gain (over impulse imaging) of $9.6$ dB at low light conditions, $2)$ Amongst the motion deblurring systems, motion invariant system provides the best performance with SNR gain of $7.5$ dB at low light conditions, and $3)$ Lytro provides the best performance amongst compared light field systems with SNR gain of $12$ dB at high light conditions. 

While the results reported in this paper are specific to EDOF, motion deblurring and light field cameras, the framework can be applied to analyze any linear CI camera. In the future, we would like to use our framework to learn priors and analyze multiplexing performance for other types of datasets (e.g. videos, hyperspectral volumes, reflectance fields). Of particular interest is the analysis of compressive CI techniques. Analyzing the performance of compressed sensing matrices has been a notoriously difficult problem, except in a few special cases (e.g. Gaussian, Bernouli, and Fourier matrices). Our framework can gracefully handle any arbitrary multiplexing matrix, and thus could prove to be a significant contribution to the compressed sensing community. By the same token, we would like to apply our analysis to overdetermined systems so that we may also analyze multiple image capture CI techniques (e.g. Hasinoff et al. \cite{Hasinoff:09} and Zhang et al. \cite{zhang2010defocusdenoise}). Finally, and perhaps most significantly, we would like to apply our framework towards the problem of parameter optimization for different CI techniques. For instance, we may use our framework to determine the optimal aperture size for focal sweep cameras, the optimal flutter shutter code for motion deblurring, or the optimal measurement matrix for a compressed sensing system. In this way, we believe our framework can be used to exhaustively analyze the field of CI research and provide invaluable answers to existing open questions in the field.

%% file: authorBio.tex
\begin{IEEEbiography}[{\includegraphics[width=1in,height=1.25in,clip,
keepaspectratio] {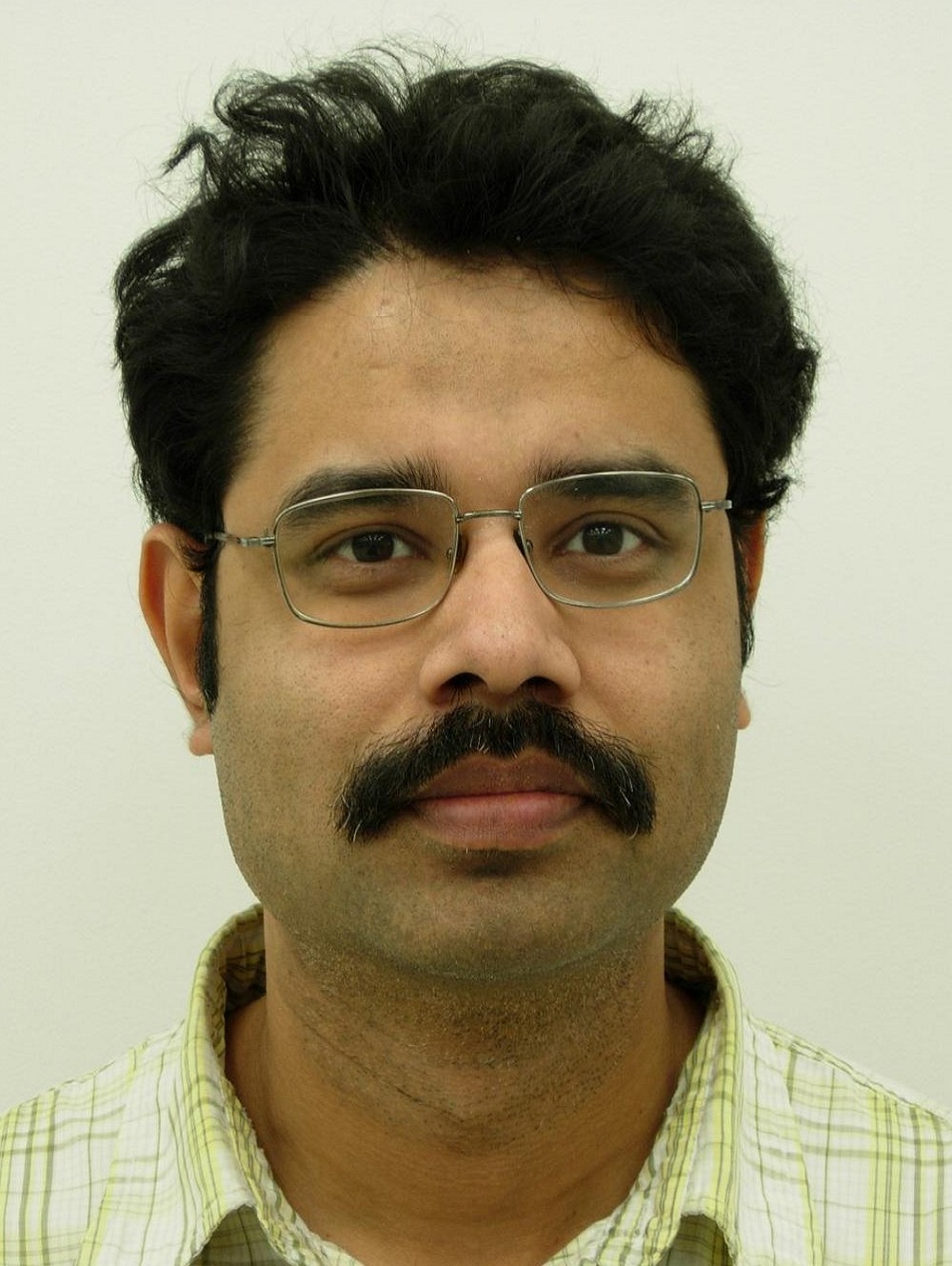}}]{Kaushik Mitra} is currently a postdoctoral research associate in the
Electrical and Computer Engineering department of Rice University. His research interests are in computational imaging,
computer vision and statistical signal processing. He earned his
Ph.D. in Electrical and Computer Engineering from the University of Maryland, College Park, where his research focus was
on the development of statistical models and optimization algorithms for computer vision problems
\end{IEEEbiography}

\begin{IEEEbiography}[{\includegraphics[width=1in,height=1.25in,clip,
keepaspectratio] {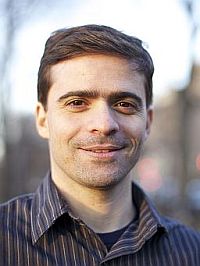}}]{Oliver Cossairt} is an Assistant Professor at Northwestern University.
His research interests lie at the intersection of optics, computer
vision, and computer graphics. He earned his PhD. in Computer Science
from Columbia University, where his research focused on the design and
analysis of computational imaging systems. He earned his M.S. from the
MIT Media Lab, where his research focused on holography and
computational displays. Oliver was awarded a NSF Graduate Research
Fellowship, Best Paper Award at ICCP 2010, and his research was
featured in the March 2011 issue of Scientific American Magazine.
Oliver has authored ten patents on various topics in computational
imaging and displays. 
\end{IEEEbiography}

\begin{IEEEbiography}[{\includegraphics[width=1in,height=1.25in,clip,
keepaspectratio] {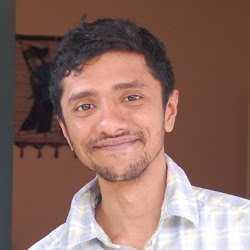}}]{Ashok Veeraraghavan} is currently an Assistant  Professor of Electrical and Computer Engineering at Rice University, Tx, USA. At Rice University, Prof. Veeraraghavan directs the Computational Imaging and Vision Lab. His research interests are broadly in the areas of computational imaging, computer vision and robotics. Before joining Rice University, he spent three wonderful and fun-filled years as a Research Scientist at Mitsubishi Electric Research Labs in Cambridge, MA. He received his Bachelors in Electrical Engineering from the Indian Institute of Technology, Madras in 2002 and M.S and PhD. degrees from the Department of Electrical and Computer Engineering at the University of Maryland, College Park in 2004 and 2008 respectively. His thesis received the Doctoral Dissertation award from the Department of Electrical and Computer Engineering at the University of Maryland. 
\end{IEEEbiography}